\documentclass[11pt,a4paper]{article}
\usepackage{times,latexsym}
\usepackage{url}
\usepackage[T1]{fontenc}
\usepackage{microtype}
\usepackage{enumitem}
\usepackage{enumitem}
\setenumerate[1]{itemsep=0pt,partopsep=0pt,parsep=\parskip,topsep=5pt}
\setitemize[1]{itemsep=0pt,partopsep=0pt,parsep=\parskip,topsep=5pt}
\setdescription{itemsep=0pt,partopsep=0pt,parsep=\parskip,topsep=5pt}
\usepackage{amsthm,amsmath,amssymb}
\usepackage{mathrsfs}
\usepackage{graphicx}
\usepackage{subfigure}
\usepackage{booktabs}
\usepackage{multirow}
\usepackage{CJKutf8}
\usepackage{makecell}
\usepackage{color}
\usepackage[acceptedWithA]{tacl2021v1}
\usepackage{tacl2021v1}
\usepackage{xspace,mfirstuc,tabulary}

\newif\iftaclinstructions
\taclinstructionsfalse % AUTHORS: do NOT set this to true
\iftaclinstructions
\renewcommand{\confidential}{}
\renewcommand{\anonsubtext}{(No author info supplied here, for consistency with
TACL-submission anonymization requirements)}
\newcommand{\instr}
\fi

\iftaclpubformat % this "if" is set by the choice of options

\else

\fi

\title{Link the World: Improving Open-domain Conversation with \\ Dynamic Spatiotemporal-aware Knowledge}

\author{Han Zhou, Xinchao Xu, Wenquan Wu, Zheng-Yu Niu, Hua Wu \\ {\bf Siqi Bao, Fan Wang, Haifeng Wang} \\ 
Baidu Inc., Beijing, China \\
\texttt{\{zhouhan05, xuxinchao, wuwenquan01\}@baidu.com}
}

\date{}

\begin{document}
\maketitle
\begin{abstract}
Making chatbots world aware in a conversation like a human is a crucial challenge, where the world may contain dynamic knowledge and spatiotemporal state. Several recent advances have tried to link the dialog system to a static knowledge base or search engine, but they do not contain all the world information needed for conversations. In contrast, we propose a new method to improve the dialogue system using spatiotemporal aware dynamic knowledge. We utilize service information as a way for the dialogue system to link the world. The system actively builds a request according to the dialog context and spatiotemporal state to get service information and then generates world aware responses. To implement this method, we collect DuSinc, an open-domain human-human dialogue dataset, where a participant can access the service to get the information needed for dialogue responses. Through automatic and human evaluations, we found that service information significantly improves the consistency, informativeness, factuality, and engagingness of the dialogue system, making it behave more like a human. Compared to the pre-trained models without spatiotemporal aware dynamic knowledge, the overall session-level score was improved by 60.87\%. The collection dataset and methods will be open-sourced.

% Making chatbots to perceive the world in a conversation like a human is a crucial challenge,  however their world knowledge and state are frozen in pre-trained models during training. Several recent advances have tried to search Internet pages in a conversation for more comprehensive knowledge, but the state of the chat is ignored, such as time, location. Building on these considerations, we propose a new method for chatbots to link the world. Specifically, we use the information-rich service APIs with spatio-temporal awareness as the external information source to update statements of chatbot. The system actively requests service information and uses it for response generation during the dialog. To implement this method, we collect DuSinc, the first open-domain service-information augmented human-human dialogue dataset. On this basis, we developed a benchmark system. Through automatic and human evaluations, we found that linking the world through service information significantly improved the multi-dimensional capabilities of chatbots to behave more like humans. The session-level overall score in human evaluation is improved by 59.29\% compared with the initial pre-training model, and the engaging score in each turn is improved by 46.82\%. Our collection data and methods will be open-sourced.
\end{abstract}

\begin{figure}[!t]
    \centering
    \includegraphics[width=8cm]{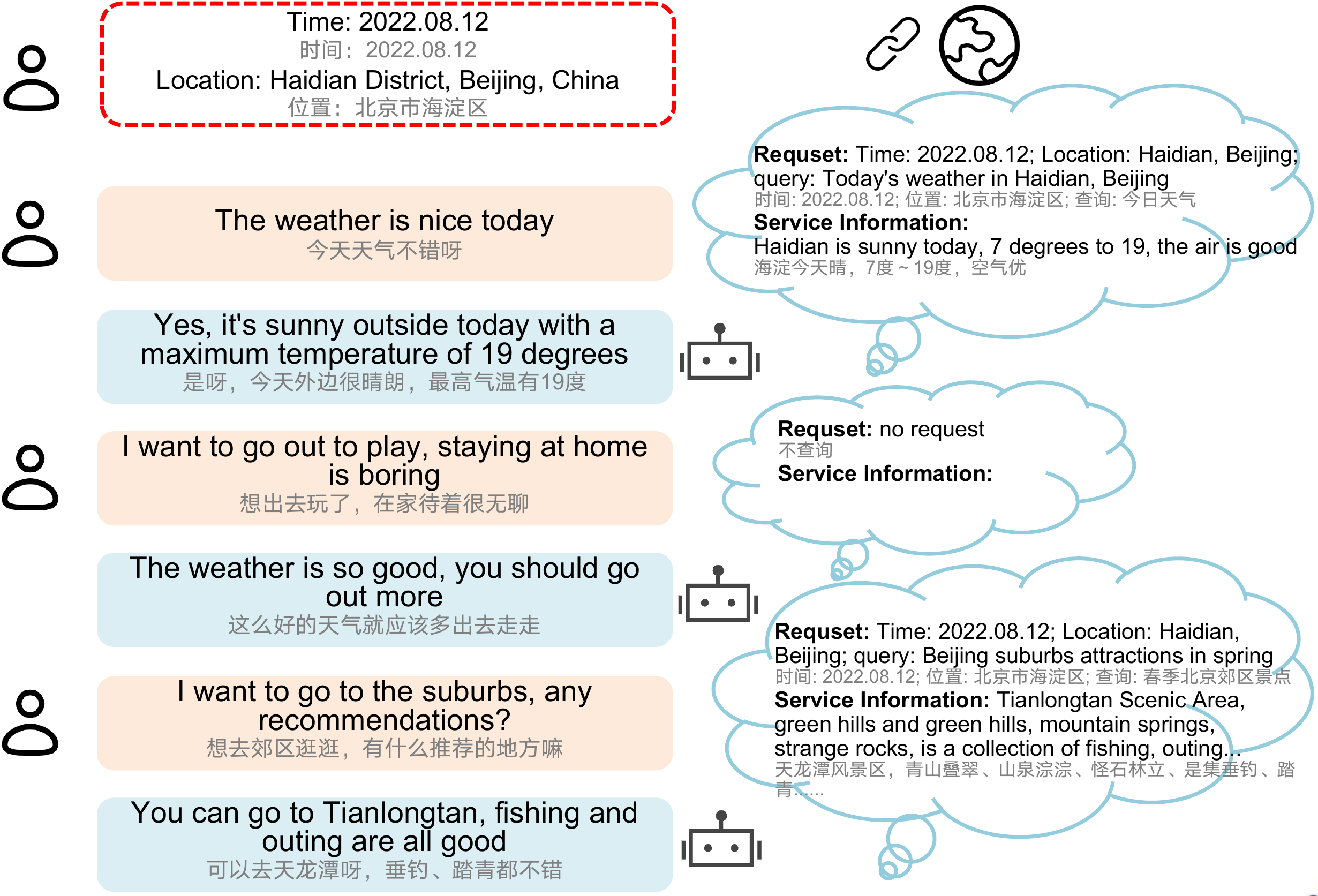}
    \caption{An example of using service information to link the world, the dialogue system accesses the service system to get relevant knowledge based on the current time and the user's location, and uses it to generate responses.}
    \label{fig:intro}
\end{figure}

\section{Introduction}
Benefiting from the recent progress in the large-scale language model, the level of multi-turn dialog system is approaching the human \cite{roller2021recipes}. It is well known that standard large language models can fluently generate responses but cannot update knowledge in time, and it is not easy to perceive the spatiotemporal state of the world \cite{hallucination_conversation}. It can only use the knowledge frozen in the model weights and cannot pay attention to the external world in a dialog like human beings \cite{lazaridou2021mind}. In other words, humans not only rely on their own knowledge (similar to the model weights), but also on the world in which interlocutors are located. Therefore, linking the world is the key to making dialog systems like a human or even beyond a human.

% Unfortunately, aggregating information from static state knowledge is a compromise method that may result in incorporating parts of mismatch statements into one factually incorrect response \cite{bb2, DBLP:journals/corr/abs-2107-07566, izacard2020distilling}. Actually, it can only use the knowledge frozen in the model weights and cannot pay attention to the external world in dialog like human beings \cite{lazaridou2021mind}. In other words, humans not only rely on their own background (similar to the model weights) but also on the world in which interlocutors are located. Therefore, linking the world to update their statements is the key to making the chatbot more human-like.

The external world that the dialogue system pays attention to contains dynamic knowledge and spatiotemporal state (\citealp{hagoort2004integration}; \citealp{hoffart2011yago2}; \citealp{venhuizen2019expectation}). Several recent advances only focused on how to construct world knowledge. They attempt to introduce external information sources into the dialogue system to supplement the inside, such as text \cite{dinan2018wizard}, QA \cite{lee-etal-2019-latent-copy}, and graphs \cite{adolphs2021reason}. However, these static knowledge bases are difficult to update promptly and cannot cover extensive topics. Wizard of Internet \cite{komeili2022internet} tries to utilize search engines as dynamic knowledge sources, improving timeliness and knowledge. Still, complex web pages seem challenging to be loaded directly by language models \cite{shuster2022language}.
Meanwhile, world knowledge should also include common skills. LAMDA is an initial attempt to introduce calculator and translation skills into the dialogs \cite{thoppilan2022lamda}. Above all, none of those methods can incorporate a spatiotemporal state, which is significant for building more human-like dialog systems.

% The world environment can contain two parts, knowledge and state (\citealp{hagoort2004integration}; \citealp{hoffart2011yago2}; \citealp{venhuizen2019expectation}). In previous work, much work focused on how to construct world knowledge. Large-scale language models learn the world knowledge contained in the text, and some works try to introduce external information sources into the dialogue system to supplement the inside, such as text, QA, and graphs. However, these static knowledge bases are difficult to update in time and cannot cover broad domains. Wizard of Internet tries to use search engines as knowledge sources, which have improved timeliness and richness. Still, complex web pages seem challenging to use directly by language models. At the same time, few works consider the world state in dialogue, which includes spatio-temporal perception, dialogue scene perception, etc. This is also important for building more human-like dialogue systems.

Based on these considerations, we use the service information to obtain external information required by the dialogue system, similar to how humans link world knowledge through various service APIs. Service Information has the following characteristics: 1) Dynamic and spatiotemporal-aware, the resources linked by service APIs are dynamically changing, including highly timely information and location-based services; 2) Various resources, service information not only includes search engines, personalized recommendations, question answering system, and also supports skills such as calculator, joke telling, stock query, and even multi-modal resources; 3) Concise, different from the web page, the information returned is often condensed paragraphs, which will be more helpful for language model understanding. In this paper, we aggregate multiple industrial-grade service APIs into an information source to which the dialog system can link.

In order to make language models use service information, we collect \textbf{DuSinc}, a \textbf{S}ervice-\textbf{in}formation augmented open-domain \textbf{c}onversation dataset in Chinese. The two annotators play the roles of USER and BOT, respectively; USER starts discussions based on their specific interests and whereabouts. Based on the current dialog context and the spatiotemporal state of the user, the BOT must determine whether more world information is needed. If needed, it constructs a request to get the service information and then provides a response based on that information; otherwise, it responds directly.

% In order for the language model to have the ability to link the world, the model takes the chat time and the user's geographic location as input, actively generates a request at the right time to access the service APIs, and generates a response based on the returned information and dialog context. To train and evaluate the capabilities of the model, we collect DuSinc, a service-information augmented Chinese open-domain dialogue dataset. Two annotators perform the roles of USER and BOT, respectively; USER starts discussions depending on their specific interests and whereabouts. Based on the present conversation context and world state, the BOT must determine whether more world information is required. If required, it will construct a request to get the service's information using the APIs and then provide a response containing the information; otherwise, it will respond directly.

Furthermore, we develop a benchmark that consists of two tasks query generation and response generation based on service information. We demonstrate that using dynamic spatiotemporal awareness information can improve the dialogue ability of different pre-trained models. Our method is general and performs well even on unseen topics, and it shows greater advantages over dialogue systems that use web page knowledge. We also analyze the model parameter scale and the impact of different training methods. Both automatic metrics and human evaluations show that our method has significant consistency, informativeness, factuality, and engagingness. Compared with the methods without linking the world, the overall session-level indicators are improved by 60.87\%.

% Furthermore, we develop a benchmark model that simulates the behavior of the BOT in DuSinc to automatically generate requests, access world information through service APIs, and generate responses. We verify that linking the world can help dialogue systems further improve performance on different pre-trained language models. We also analyze in detail the effect of the parameter scale of the pre-trained model on the ability to use the service information. At the same time, we consider the influence of different types of external knowledge sources as world information. Both automatic metric evaluation and human-machine evaluation show that our proposed system significantly improves the human-like abilities of open-domain dialogue in terms of information richness, factual accuracy, and more. At the same time, the ability is general and also performs well in unseen conversation topics.

In summary, this paper makes the following contributions:
\begin{enumerate}
\item We propose a method to link the world through service information so that the dialog system can utilize dynamic knowledge and spatiotemporal state awareness.
\item We collect DuSinc; the first open-domain Chinese service information augmented human-human dialogue dataset. It contains more than 2,500 topics, close to 10,000 conversations, and 50,000 service requests. This dataset will be open-sourced soon;
\item We build strong benchmarks on the DuSinc dataset using state-of-the-art large-scale pre-trained models for both query generation and response generation. Extensive experiments demonstrate that linking the world through service information can significantly improve conversational performance.
\end{enumerate}

\begin{table*}[!htb]
\small
\renewcommand\arraystretch{1}
\centering
\tabcolsep=0.1cm
\begin{tabular}{ccccccccc}
\toprule
\textbf{Dataset}       & \textbf{\begin{tabular}[c]{@{}c@{}}Know.\\ Type\end{tabular}} & \textbf{\begin{tabular}[c]{@{}c@{}}Time\\ Sensitive\end{tabular}} & \textbf{\begin{tabular}[c]{@{}c@{}}Location\\ Sensitive\end{tabular}} & \textbf{\begin{tabular}[c]{@{}c@{}}Skills\\ Know.\end{tabular}} & \textbf{Domain} & \textbf{Lang.} & \textbf{\begin{tabular}[c]{@{}c@{}}\# chars \\ per uttr\end{tabular}} & \textbf{\# uttr} \\ \midrule
\textbf{CMU DoG}       & Text                                                          & $\times$                                                                & $\times$                                                                    & $\times$                                                              & 1               & en             & 11.8                                                                  & 130K             \\
\textbf{OpenDialKG}     & Graph                                                         & $\times$                                                                & $\times$                                                                    & $\times$                                                              & 4               & en             & -                                                                     & 91K              \\
\textbf{Topcial Chat}  & Text                                                          & $\times$                                                                & $\times$                                                                    & $\times$                                                              & Open            & en             & 19.6                                                                  & 235K             \\
\textbf{WoW}        & Text                                                          & $\times$                                                                & $\times$                                                                    & $\times$                                                              & Open            & en             & -                                                                     & 202K             \\
\textbf{WizInt}         & Web                                                      & \checkmark                                                               & $\times$                                                                    & $\times$                                                              & Open            & en             & 19.1                                                                  & 94K              \\ \midrule
\textbf{DuConv}        & Text \& Graph                                                 & $\times$                                                                & $\times$                                                                    & $\times$                                                              & 1               & zh             & 10.6                                                                  & 270K             \\
\textbf{KdConv}         & Text \& Graph                                                 & $\times$                                                                & $\times$                                                                    & $\times$                                                              & 3               & zh             & 20.8                                                                  & 86K              \\ \midrule
\textbf{DuSinc (ours)} & Service                                                   & \checkmark                                                              & \checkmark                                                                   & \checkmark                                                             & Open            & zh             & 21.7                                                                  & 102K             \\ \bottomrule
\end{tabular}
\caption{Comparison between DuSinc and other human-labeled knowledge-grounded dialogue datasets.}
\label{tab:Comparison}
\end{table*}

\section{DuSinc Collection}
This section mainly introduces the collection of the DuSinc dataset and the service information. We considered the following setting for data collection: two participants chat around a topic, which can change naturally during the conversation. There is an information asymmetry between the two participants. One of them plays the role of USER, who is set in a specific geographic location and has a topic of interest; the other participant plays the role of BOT and can use service information in the conversation. The conversation is initiated by the USER, and both sides speak alternately.

\subsection{Service Information}
The service API is a real-time updated source of information for chatbots to link the world. It accepts a request that consists of a query and an spatiotemporal state, and then delivers accurate relevant information. Specifically, we have developed a dynamically updated, real-time accessible system, which is an aggregation of many industrial-grade information sources. These include industrial-grade end-to-end deep Q\&A services, location-based personalized recommendation services (such as  weather of today, nearby western restaurants suitable for eating with children, driving routes to Beijing Railway Station, etc.), skill-based services (such as calculators, perpetual calendar, translation, stock price inquiry, etc.). The returned information is sensitive to spatiotemporal circumstances and could satisfy the external knowledge needs of open domain chat in a comprehensive manner. The system serves both data collection and model deployment inference.
% Service can be considered a black-box system that takes a request consisting of a query and environment state and returns precise paragraphs of text. In terms of the specific implementation, we have developed a dynamically updated, real-time accessible system, which is an aggregation of numerous APIs. These include industrial-grade end-to-end deep Q\&A services, location-based personalized recommendation services (such as today's weather, nearby western restaurants suitable for eating with children, driving routes to Beijing Railway Station, etc.), skill-based services (such as calculators, perpetual calendar, translation, stock price inquiry, etc.). In this system, requests are serially passed through various APIs in the order of skill services, location-based services, and Q\&A services until a result is returned. The returned information is sensitive to spatiotemporal scenarios and can broadly cover the external knowledge requirements of open domain chat. The system is used for both data collection and model deployment inference.

\subsection{USER Settings}
\begin{figure}[!h]
    \centering
    \includegraphics[width=8cm]{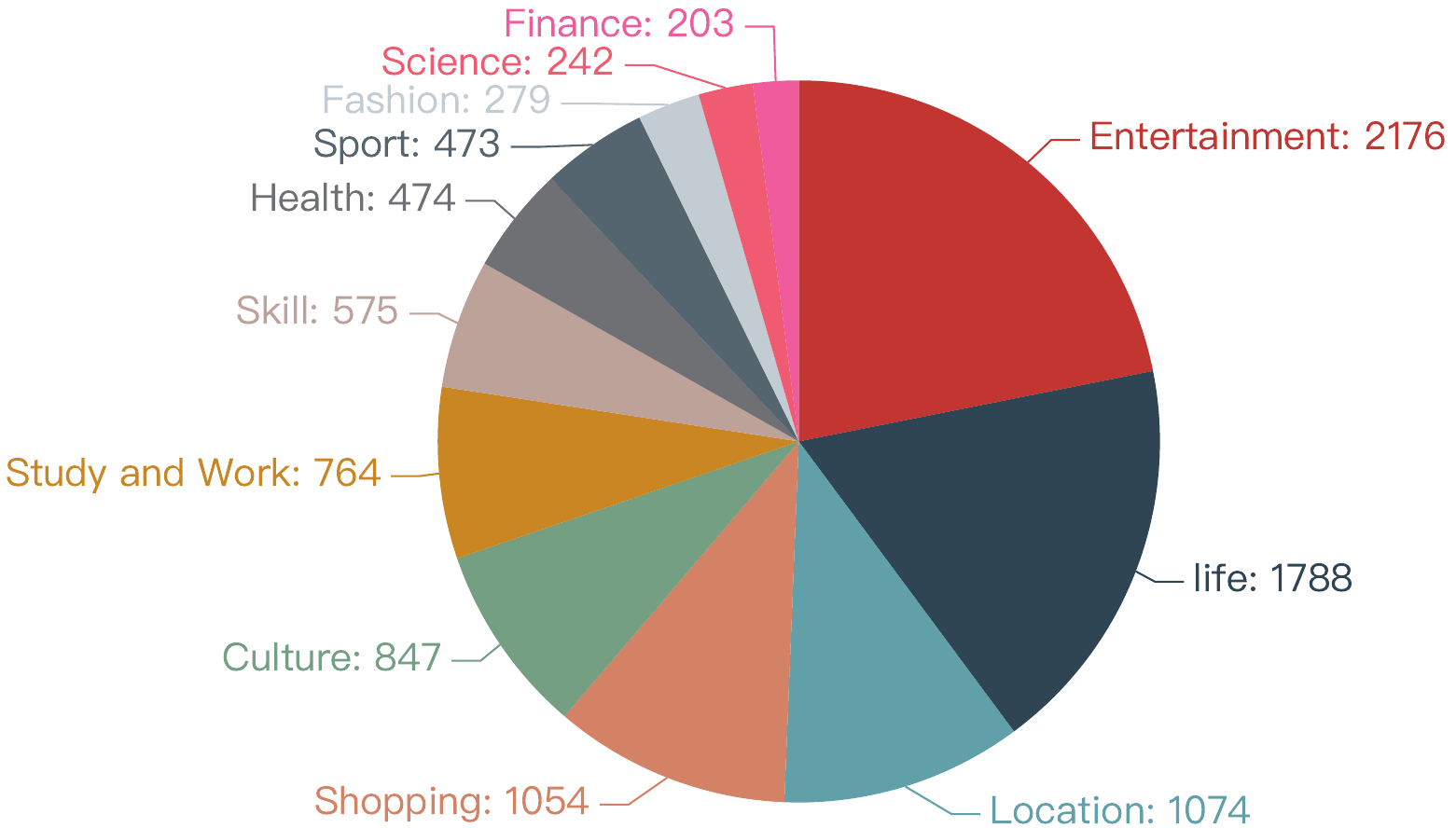}
    \caption{First-level topic distribution in DuSinc dataset.}
    \label{fig:topical}
\end{figure}

A participant playing USER should comply with: \emph{I am a person in a specific location, and I want to chat with each other around a topic that interests me.} 
Before starting the conversation, USER will be randomly assigned a geographic location with latitude and longitude (a city from China), and they are allowed to have a conversation based on their geographic location information, such as \emph{What's the weather like today?}, \emph{What's the food nearby?}. 
To engage in more in-depth discussions on this topic, participants must pick the subtopic that most interests them.
We have created a three-tiered topic system. The first-level categories are fixed and cannot be altered; they include 12 sorts such as \emph{life}, \emph{sports}, \emph{technology}, etc. Figure \ref{fig:topical} shows the distribution of first-level topics, where users are most interested in entertainment, life, and local services.
The second-level categories are preset but can be modified or supplemented, such as \emph{life-fishing}, \emph{sports-badminton}, etc. 
The third-level category, which needs to be supplemented by participants, is a further refinement of the topic, which determines that our dialogue domain is open, such as \emph{sports-badminton-Lin Dan}.

\begin{table*}[!htb]
\small
\setlength\tabcolsep{6pt}
\centering
\begin{tabular}{cccccc}
\toprule
\textbf{DuSinc Statistics}              & \textbf{Train} & \textbf{Valid} & \textbf{Seen Test} & \textbf{Unseen Test} & \textbf{Total} \\ \midrule
\textbf{\# dialogs}                     & 8,449          & 500            & 500                & 500                  & 9,949          \\
\textbf{\# utterances}                  & 86,765         & 5,117          & 5,159              & 5,106                & 102,147        \\
\textbf{Avg. \# chars per USER uttr}    & 15.44          & 15.26          & 15.19              & 15.93                & 15.44          \\
\textbf{Avg. \# chars per BOT uttr}     & 24.67          & 24.47          & 25.01              & 25.55                & 24.72          \\
\textbf{Avg. \# chars per query}        & 6.33           & 6.31           & 6.19               & 6.96                 & 6.35           \\
\textbf{Avg. \# chars per service text} & 317.19         & 338.59         & 323.69             & 300.80               & 317.78         \\
\textbf{Avg. \# service turn percent}   & 52.39\%        & 51.81\%        & 52.63\%            & 50.98\%              & 52.30\%        \\
\textbf{Avg. \# other service}          & 0.24           & 0.20           & 0.26               & 0.24                 & 0.24           \\
\textbf{\# topics level 1/2/3}          & 12/339/2,055   & 12/77/219      & 12/76/216          & 10/12/127            & 12/378/2,361   \\
\textbf{\# locations}                   & 416            & 297            & 297                & 298                  & 416            \\ \bottomrule
\end{tabular}
\caption{Statistics of DuSinc.}
\label{tab:Statistics}
\end{table*}

\subsection{BOT Settings}
A participant playing BOT should comply with: \emph{I need to have a coherent dialogue with the USER and judge whether replying to the current context requires external knowledge, if necessary, access to the service system, and compose a response based on the service information}. The topic and location of interest to the USER are visible to the BOT player. When BOT uses the service system, the query in request written by BOT should be specific short sentences, rather than simply copying the keywords or entities mentioned in the context. This will make the queried knowledge more detailed and suitable for the current context. In addition, we believe that the actual dialogue scene does not need to use external knowledge all the time, so in the dataset collection, we access the state-of-the-art dialogue model PLATO-XL auxiliary annotation. BOT can see the responses generated by the current turns of PLATO-XL \cite{bao2021plato}. If the generated responses already meet the requirements, they can make adjustments on this basis without using external services.

\subsection{Quality Control}
To assure the quality of the collected data, we have meticulously optimized the annotators, tools, specifications, and processes.
\begin{itemize}[leftmargin=*]
\item The annotators are volunteers recruited from Chinese universities, which increases the variety and quality of the conversation.
\item We build a customized annotation tool to improve annotation efficiency and specification. It supports online chat, role assignment, user information configuration, and the service system. 
\item For dialog coherence, responses participants should be colloquial rather than copying information. 
\item Each discussion group has five turns and at least two in-depth turns utilizing the service system. Hello, farewell, etc., are banned since they hinder our work.
\item BOT may repeatedly ask for external services if dissatisfied. Even if not utilized, requests and knowledge are recorded. 
\item The USER rates the dialog quality of BOT after the interaction.
\end{itemize}
More quality control strategies are discussed in Appendix \ref{sec:data}.

\subsection{Dataset Statistics}
As demonstrated in Table \ref{tab:Comparison}, DuSinc is the first open-domain Chinese knowledge-grounded dialogue dataset containing 2,361 topics. More information on topic distribution is in Appendix \ref{sec:data}. Meanwhile, DuSinc is the first dialog dataset to use the dynamic service information and is spatiotemporal aware, while it contains more skill knowledge. And the average number of characters per utterance reaches 21.7, which means the conversation is informative. We calculate Distinct-2 \cite{li2016diversity}, the diversity metric for users and bots, respectively. The value of DuSinc is 0.32/0.39, which is only 0.13/0.22 in DuConv \cite{wu2019proactive} and 0.24/0.23 in KdConv \cite{zhou2020kdconv}. As can be seen, the dialogue in DuSinc is more diverse.

\begin{figure*}[!h]
    \centering
    \includegraphics[width=15cm]{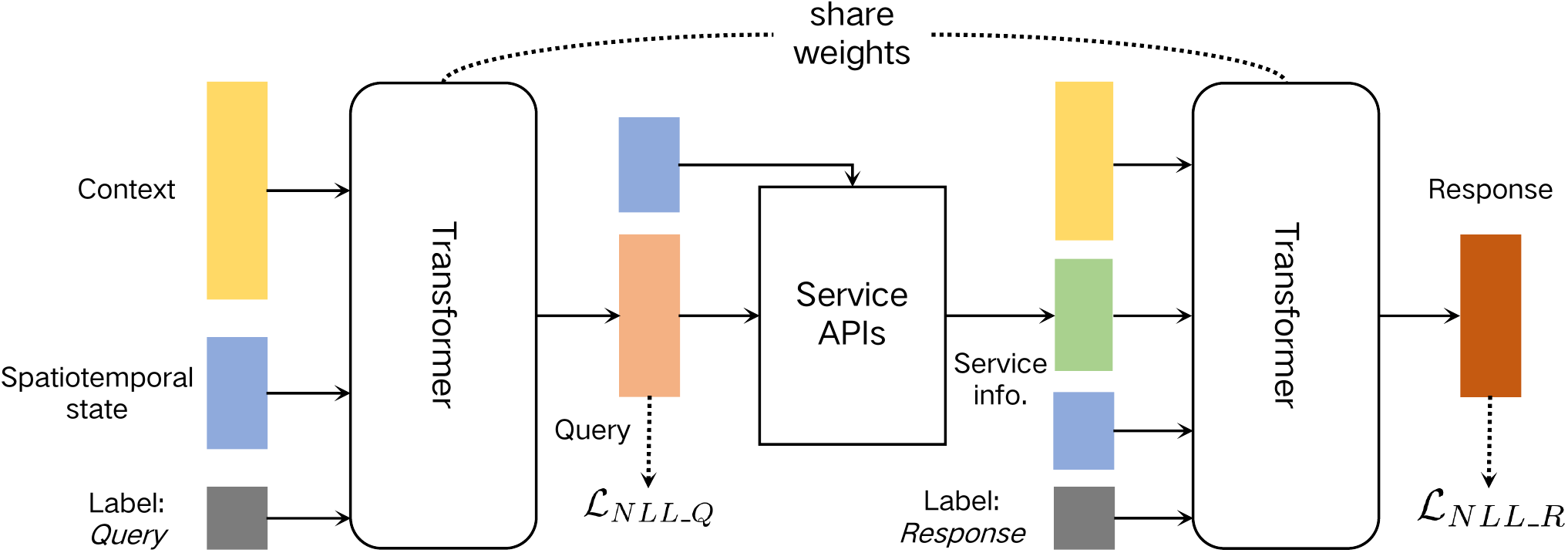}
    \caption{Knowledge dialogue model structure for service information enhancement.}
    \label{fig:model}
\end{figure*}

In total, more than 300 volunteers participated in data collection. As shown in Table \ref{tab:Statistics}, the overall collected data includes 9,949 sets of dialogs with 102,147 utterances, which are divided into 86,765 utterances for training, 5,117 utterances for validation, and 5,159 utterances for seen testing, and 5,106 utterances for unseen testing. The second-level topics in the test set have never appeared in the training set. 52.30\% of BOT turns used external knowledge, indicating that it integrates chitchat and knowledge-grounded conversations, which we believe is consistent with the knowledge distribution in real conversations. The average character in the query is 6.35, which indicates that in most cases, the query is a specific short utterance, not just a keyword, which is the key to querying accurate service information. The service text is a paragraph with an average length of 317.78 characters, which is different from a long-form webpage or a graph triple with less information. In knowledge-used turns, there were an average of 0.24 requests for services but no use. These behaviors may help to build more appropriate requests.

\section{Method}
We develop a model that is capable of incorporating information from external services. The system generates a response $R=b_t$ given the USER information $M$ and multi-turns of context $C=\{u_1,b_1,\ldots,u_{t-1},b_{t-1},u_t\}$. Specifically, as shown in Figure \ref{fig:model}, the model first generates the query $Q=q_t$ according to $C$ and the spatiotemporal state $S$, and uses the request composed of $S$ and $Q$ to access the service system, gets relevant knowledge $K=k_t$, and further adds the $K$ to generate a response. Among them, $u_t$ denotes the utterance of the USER turn $t$, $b_t$ denotes the utterance of the BOT turn $t$. The system is used to simulate the actions of the BOT.

We divide the system into two stages: request construction and response generation. We will introduce the details of these two modules in the following subsections.

\subsection{Request Construction}
The request contains spatiotemporal state and query. The state can be obtained automatically, so the functions of this module include judging whether to use external knowledge and generating a query. We fine-tune the pre-trained model to generate a query and a particular piece of text "no request" when no external knowledge is required. The input to the model is the sum of the corresponding token, type, and position embeddings of $S$ and $C$. During training, we minimize the following negative log-likelihood (NLL) loss:\vspace{-0.05cm}
\begin{small}
\begin{equation}
\begin{split}
\mathcal{L}_{NLL\_Q} &= -\mathbb{E}\ {\rm log}\;p(Q|S,C) \\
&=-\mathbb{E}\ \sum_{j=1}^{|Q|}\ {\rm log}\;p(q_j|S,C,Q_{<j})
\end{split}
\end{equation}\vspace{-0.1cm}
\end{small}
Where $q_{<j}$ denotes previously generated tokens in query $Q$. Then, we use the request to access the service system and get the relevant service knowledge $K$.

\subsection{Response Generation}
Similarly, we fine-tune the pre-trained model to generate $R$, given $S$, $C$, and $K$ (if any). Different input parts are distinguished using typed embeddings. We minimize the following NLL loss:\vspace{-0.2cm}
\begin{small}
\begin{equation}
\begin{split}
\mathcal{L}_{NLL\_R} &= -\mathbb{E}\ {\rm log}\;p(R|S,C,K) \\
&=-\mathbb{E}\ \sum_{j=1}^{|R|}\ {\rm log}\;p(r_j|S,C,K,R_{<j})
\end{split}
\end{equation}\vspace{-0.1cm}
\end{small}
$r_{<j}$ denotes previously generated tokens in response $R$. The model weights model in the query and response generation tasks are shared during model training. We append additional task labels \emph{"Query"} or \emph{"Response"} to the input text to distinguish different tasks. Therefore, the loss function for system optimization is:\vspace{-0.3cm}
\begin{small}
\begin{equation}
\mathcal{L}_{NLL} = \mathcal{L}_{NLL\_Q} + \mathcal{L}_{NLL\_R}
\end{equation}
\end{small}

\section{Experiments}
\subsection{Evaluation Settings}

We carry out service information augmented conversation task experiments on the DuSinc dataset. These methods perform both automatic and human evaluations.

\begin{table*}[!htb]
\renewcommand\arraystretch{1.2}
\small
\centering
\begin{tabular}{cc|cc|cccc}
\toprule
\multirow{2}{*}{\textbf{Model}} & \multicolumn{1}{c|}{\multirow{2}{*}{\textbf{Parameters}}} & \multicolumn{2}{c|}{\textbf{Query Generation}}       & \multicolumn{4}{c}{\textbf{Response Generation}}                     \\ \cline{3-8} 
                                & \multicolumn{1}{c|}{}                                     & \textbf{ACC}   & \multicolumn{1}{c|}{\textbf{F1}}    & \textbf{F1}    & \textbf{KF1}   & \textbf{BLEU}  & \textbf{DIS} \\ \midrule
\multicolumn{8}{c}{\textbf{Seen Test / UnSeen Test}}                                                                                                                                                                                    \\ \midrule
\textbf{T5 (golden kg)}         & \multicolumn{1}{c|}{\multirow{2}{*}{620M}}                & 67.2/65.9          & \multicolumn{1}{c|}{49.7/48.7}          & 24.2/23.3          & 12.6/13.1          & 47.6/50.9          & 45.1/46.2             \\
\textbf{T5 (no kg)}             & \multicolumn{1}{c|}{}                                     & 62.0/59.3          & \multicolumn{1}{c|}{43,5/43.4}          & 15.4/15.5          & 7.0/7.0          & 35.2/47.7          & 26.9/25.6             \\
\textbf{BART (golden kg)}       & \multicolumn{1}{c|}{\multirow{2}{*}{580M}}                & \textbf{69.5}/\textbf{70.2} & \multicolumn{1}{c|}{52.8/54.8}          & 29.2/29.2          & \textbf{17.3}/\textbf{19.7} & 49.8/48.9 & 54.0/49.3             \\
\textbf{BART (no kg)}           & \multicolumn{1}{c|}{}                                     & 68.2/68.5          & \multicolumn{1}{c|}{49.0/44.2}          & 19.0/19.3          & 12.6/9.3          & 47.6/41.0          & 45.1/43.0             \\
\textbf{EVA2.0 (golden kg)}     & \multicolumn{1}{c|}{\multirow{2}{*}{970M}}                & 59.0/57.1          & \multicolumn{1}{c|}{39.6/38.0}          & 19.7/18.5          & 4.5/4.6          & 29.6/29.4          & 33.1/28.1             \\
\textbf{EVA2.0 (no kg)}         & \multicolumn{1}{c|}{}                                     & 58.6/57.4          & \multicolumn{1}{c|}{33.4/32.3}          & 11.0/10.6          & 2.7/2.9          & 20.5/19.1          & 65.9/65.1             \\
\textbf{PLATO (golden kg)}    & \multicolumn{1}{c|}{\multirow{2}{*}{600M}}                & 68.5/69.1          & \multicolumn{1}{c|}{\textbf{54.5}/\textbf{55.4}} & \textbf{29.8}/\textbf{30.1} & 12.9/16.2          & \textbf{52.9}/\textbf{51.3}          & \textbf{55.7}/\textbf{50.6}    \\
\textbf{PLATO (no kg)}        & \multicolumn{1}{c|}{}                                     & 65.9/66.9          & \multicolumn{1}{c|}{53.6/54.1}          & 24.9/25.0          & 6.2/6.2          & 40.1/39.7          & 43.1/38.6             \\ \midrule
\end{tabular}
\caption{Automatic evaluation results of different models on the DuSinc test set, the best scores are shown in bold.}
\label{tab:Automatic-1}
\end{table*}

% \usepackage{multirow}
\begin{table*}[!htb]
\renewcommand\arraystretch{1.2}
\small
\centering
\begin{tabular}{cc|cccc|cccc}
\toprule
\multirow{2}{*}{\textbf{Model}} & \multirow{2}{*}{\textbf{\begin{tabular}[c]{@{}c@{}}Knowledge \\ Access\end{tabular}}} & \multicolumn{4}{c|}{\textbf{Seen}}                                                                                                             & \multicolumn{4}{c}{\textbf{Unseen}}                                                                                                           \\ \cline{3-10} 
                                &                                                                                       & \multicolumn{1}{c}{\textbf{F1}} & \multicolumn{1}{c}{\textbf{KF1}} & \multicolumn{1}{c}{\textbf{BLEU}} & \multicolumn{1}{c|}{\textbf{DIS}} & \multicolumn{1}{c}{\textbf{F1}} & \multicolumn{1}{c}{\textbf{KF1}} & \multicolumn{1}{c}{\textbf{BLEU}} & \multicolumn{1}{c}{\textbf{DIS}} \\ \midrule
\textbf{PLATO}                         & \textbf{no knowledge}                                                                          & 19.9                            & 4.0                              & 31.1                                & 41.3                                & 19.7                            & 4.4                              & 31.3                                & 36.7                               \\
\textbf{PLATO-FT}                        & \textbf{no knowledge}                                                                          & 24.9                            & 5.8                              & 40.1                                & 43.1                                & 25.0                            & 6.5                              & 39.7                                & 38.6                               \\
\textbf{PLATO-FID}                  &  \textbf{Web Top 16}                                                                & 24.8                            & 8.2                              & 41.6                                & 51.0                                & 25.1                            & 8.1                              & 40.2                                & 45.8                               \\
\textbf{PLATO-SINC}                      &  \textbf{Web Top 1}                                                                 & 25.8                            & \textbf{11.1}                             & 42.7                                & 51.9                                & 25.7                            & \textbf{12.4}                             & 40.2                                & 47.3                               \\
\textbf{PLATO-SINC}                      &  \textbf{Service}                                                                       & \textbf{28.1}                            & 8.4                              & \textbf{45.0}                                & \textbf{54.3}                                & \textbf{28.3}                            & 10.3                             & \textbf{46.8}                               & \textbf{49.2}                               \\ \bottomrule
\end{tabular}
\caption{PLATO's automated evaluation results using different types of external knowledge, the best scores are shown in bold.}
\label{tab:Automatic-2}
\end{table*}

% \usepackage{multirow}
\begin{table*}[!htb]
\renewcommand\arraystretch{1.2}
\footnotesize
\centering
\begin{tabular}{c|ccc|ccccc}
\toprule
\multicolumn{1}{c|}{\multirow{2}{*}{\textbf{Model}}} & \multicolumn{3}{c|}{\textbf{Query Generation}}                        & \multicolumn{5}{c}{\textbf{Response Generation}}                                     \\ \cline{2-9} 
\multicolumn{1}{c|}{}                                & \textbf{ACC}   & \textbf{F1}    & \multicolumn{1}{c|}{\textbf{PPL $\downarrow$}}   & \textbf{F1}    & \textbf{KF1}   & \textbf{BLEU} & \textbf{DIS} & \textbf{PPL $\downarrow$}    \\ \midrule
\multicolumn{9}{c}{\textbf{Seen Test / UnSeen Test}}                                                                                                                                                                              \\ \midrule
\multicolumn{1}{c|}{\textbf{PLATO-Query}}            & \textbf{67.8}/\textbf{69.8} & 53.5/55.7          & \multicolumn{1}{c|}{2.69/3.01}          & /              & /              & /               & /              & /               \\
\multicolumn{1}{c|}{\textbf{PLATO-Response}}         & /              & /              & \multicolumn{1}{c|}{/}              & \textbf{30.6}/\textbf{31.2} & 15.6/18.8          & 51.8/\textbf{53.6}           & 56.5/51.0 & \textbf{11.75}/\textbf{12.7} \\
\multicolumn{1}{c|}{\textbf{PLATO-SINC}}             & 67.2/68.3          & \textbf{54.2}/\textbf{56.6} & \multicolumn{1}{c|}{\textbf{2.53/2.79}} & 30.2/30.6          & \textbf{16.2/19.1} & \textbf{52.9/53.0}  & \textbf{58.1/51.6} & 11.84/12.88 \\ \bottomrule
\end{tabular}
\caption{The impact of query generation and response generation shared parameter training.}
\end{table*}

\textbf{Automatic evaluation settings.} We consider two tasks: query generation and response generation. In Query generation, we leverage several standard metrics: Accuracy, which measures the accuracy of whether external knowledge needs to be retrieved. The F1 \cite{dinan2019wizard} value is used to evaluate the consistency between the predicted and golden queries when the golden query exists. PPL \cite{meister2021language} can determine the coherence of the predicted query to a certain extent. In response generation, we additionally used BLEU-1 \cite{chen2014systematic} to evaluate the consistency of predicted responses with standard responses, KF1 \cite{dinan2019wizard} to evaluate the relevance of responses to knowledge, and Distinct-2 \cite{li2016diversity} to evaluate the diversity of responses in the test set \cite{li-etal-2016-diversity}. We experiment on multiple Chinese pre-trained language models, including T5 \cite{xue2021mt5}, BART \cite{liu2020multilingual}, EVA2.0 \cite{gu2022eva2}, and PLATO \cite{bao2022plato}. In addition, we investigate the performance of the model at various parameter scales and the differentiation between various categories of knowledge as world information. 
% More hyperparameter settings for model training and inference are in Appendix \ref{sec:model}.

\begin{figure*}[t]
    \centering
    \subfigure[Query PPL]{
    \includegraphics[width=4cm]{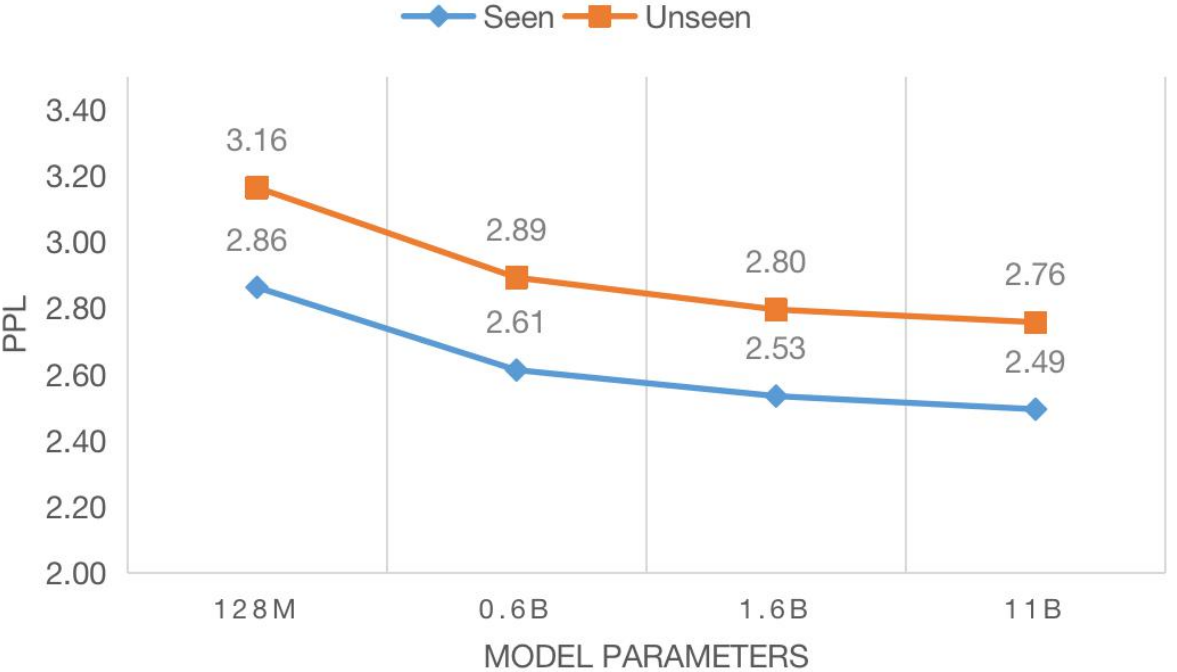}
    }\subfigure[Response PPL]{
    \includegraphics[width=4cm]{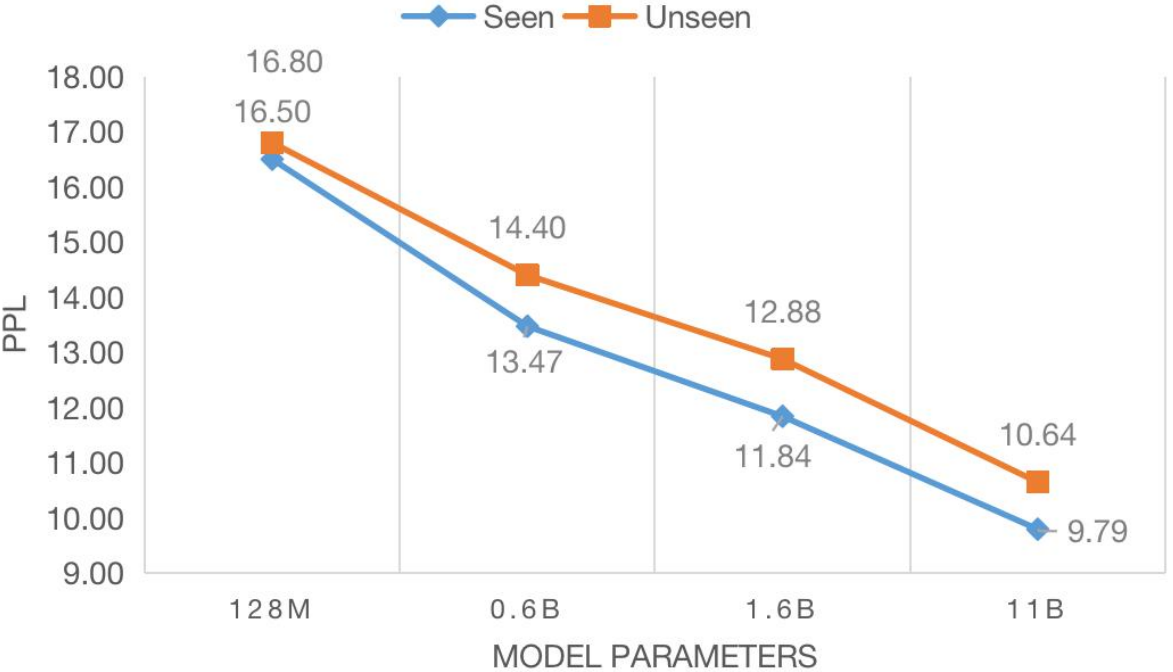}
    }\subfigure[Query F1]{
    \includegraphics[width=4cm]{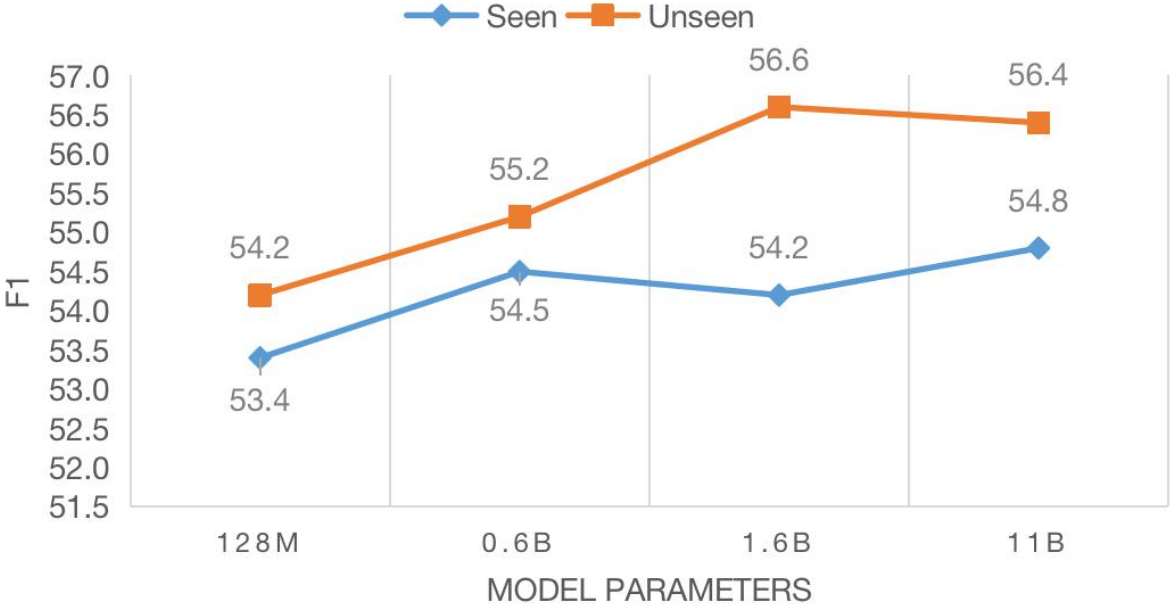}
    }\subfigure[Response F1]{
    \includegraphics[width=4cm]{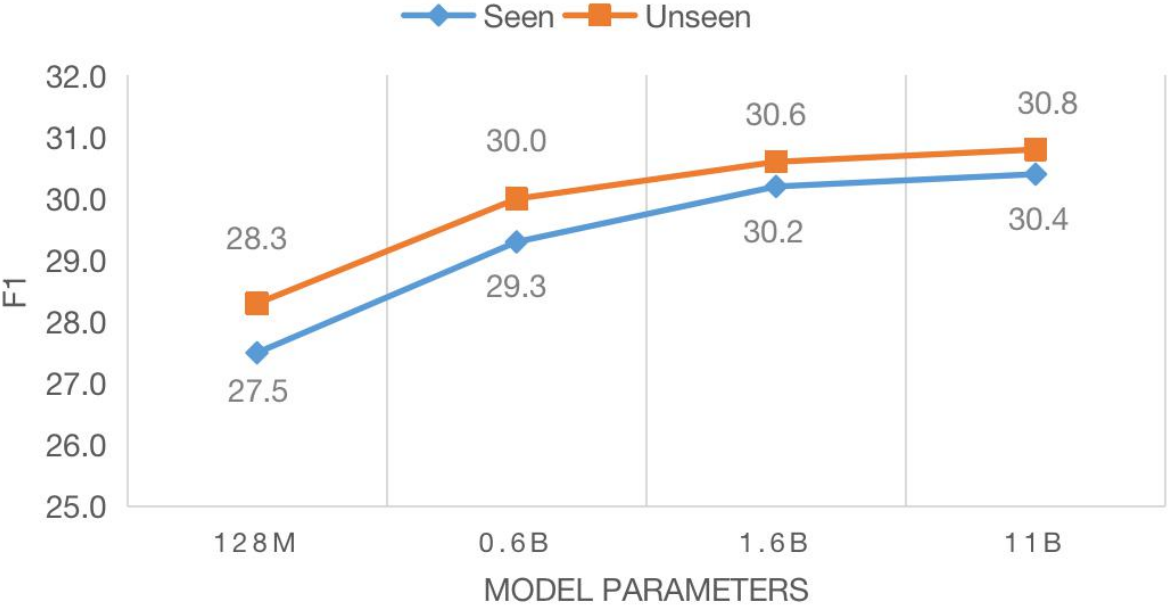}
    }
    \caption{The influence of pre-trained PLATO models with different parameter scales on the service information augmented conversation task.}
    \label{fig:parameter}
\end{figure*}

\textbf{Human evaluation settings.} We collect multi-turn human-machine dialogue for human evaluations on competitive methods in automatic evaluation. To comprehensively evaluate the ability of the methods to utilize the world information, we selected 60 topics as the initial sentences of the conversations, which can be divided into chitchat topics, in-depth topics, and spatiotemporal \& skill topics. The annotator initiates the conversations according to a given topic, and each conversation lasts at least five turns. Additional annotators score the collected dialogues at turn level and session level. Among them, at the turn level, we mainly focus on the four metrics which score is 0 or 1. 1) Consistency, whether the reply is coherent with the context and the reply itself. 2) Knowledgeable, whether the reply contains knowledge or common sense information. 3) Factual Accuracy, whether the knowledge in the reply is wrong or contrary to common sense. 4) Engaging,  whether the reply is attractive and whether it is willing to continue the dialogue with the bot. The session level overall score is a comprehensive metric with a score of 0-5, where 0 means bad and 5 means perfect.

\subsection{Automatic Evaluation Results}
\textbf{Pre-training models.} As shown in Table \ref{tab:Automatic-1}, golden kg means using the DuSinc dataset to train service information augmented models and using golden knowledge for inference. No kg means using only the dialogue context when training and inference. Adding service information can significantly improve the effect of query and response generation in the four pre-training models. At the same time, we can also see that in the seen and unseen test sets, the model with service information has little difference in automatic metrics, which shows that the method is generalized and can transfer domains and topics. Experiments were conducted using a pre-training model with comparable parameters. In comparison, PLATO and BART have better performance. PLATO has more advantages in F1, BLEU, and Distinct, while BART is reflected in KF1. We believe that pre-training corpus of BART contains a lot of knowledge text, which may be closer to service knowledge so that it is more inclined to use knowledge. PLATO seems to have a more robust dialogue generation ability. EVA2.0 performs the worst among them, which we attribute to the pre-training conversation average duration data being shorter.

\textbf{Knowledge access methods.} We compare the effects of generating responses given different external information, experimenting on PLATO with 1.6 billion parameters. 
We investigated the following settings: PLATO is an untuned pre-trained model, PLATO-FT is a model fine-tuned using only conversation context, PLATO-FID \cite{izacard2021leveraging} is a model capable of fusing several bits of knowledge during decoding, and PLATO-SINC is our suggested model for service information augmentation.
For knowledge access, no knowledge means not using external information. The web is the text from web pages. We get the relevant web pages through the Chinese search engine, and select the topk paragraphs in them through the DPR \cite{karpukhin2020dense} method. Service is the service information we propose. For Web and Service,  we use the same query predicted by the model.

As demonstrated in Table \ref{tab:Automatic-2}, incorporating external information may enhance the capabilities of various conversation system components. Service information is more favorable than web pages based on most automated criteria, particularly the consistency and diversity metrics F1 and DIS.
The KF1 metric among them is pretty low. We consider it since the length of the online text is often more significant than that of the service information, which is typically shorter.

\textbf{Share weights.} We verify that the shared weights between query and response generation tasks are effective on the PLATO model. We use the same pre-trained model to fine-tune these two tasks separately, and the resulting models are denoted as PLATO-Query and PLATO-Response. The experimental results are shown in Table \ref{tab:Automatic-1}. In both tasks, the shared-weight training model can approach or exceed the single training model on automatic metrics, with equal performance on the seen and unseen test sets. This shows that the learning of these two tasks can promote each other.

\textbf{Parameter scale.} By fine-tuning the PLATO model with 128M, 0.6B, 1.6B, and 11B parameters, we explore the relationship between the parameters of the dialogue generation model and the ability to use information from external services. We uniformly use the method of sharing weights between the two tasks of query generation and response generation for model training and automatically evaluate the two tasks separately. The experimental results are shown in Figure \ref{fig:parameter}. The PPL metric was positively correlated with the parameter scale in both tasks. However, on the F1 metric, the query generation of the model with 11 billion parameters is lower than that of the model with 1.6 billion parameters. Therefore, increasing the model size can improve the ability of the dialogue system to use world information, but for query generation tasks, fewer parameters can already have a good effect.

\begin{table*}[!htb]
\renewcommand\arraystretch{1.2}
\footnotesize
\centering
\begin{tabular}{ccccccc}
\toprule
\textbf{Model}      & \multicolumn{1}{c|}{\textbf{\begin{tabular}[c]{@{}c@{}}Knowledge \\ Access\end{tabular}}} & \textbf{Consistent} & \textbf{Knowledgeable} & \textbf{\begin{tabular}[c]{@{}c@{}}Factually \\ Incorrect  $\downarrow$\end{tabular}} & \multicolumn{1}{c|}{\textbf{Engaging}} & \textbf{Overall Score} \\ \midrule
\multicolumn{7}{c}{\textbf{All Topics}}                                                                                                                                                                                                                                                                    \\ \midrule
\textbf{PLATO}      & \multicolumn{1}{c|}{\textbf{no knowledge}}                                                & 75.44\%             & 86.47\%                & 14.99\%                                                                 & \multicolumn{1}{c|}{58.82\%}           & 2.53                   \\
\textbf{PLATO-SINC} & \multicolumn{1}{c|}{\textbf{web top 1}}                                                   & 83.48\%             & 97.48\%                & 13.33\%                                                                 & \multicolumn{1}{c|}{78.33\%}           & 3.52                   \\
\textbf{PLATO-SINC} & \multicolumn{1}{c|}{\textbf{service}}                                                     & \textbf{87.35\%}    & \textbf{98.30\%}       & \textbf{7.56\%}                                                         & \multicolumn{1}{c|}{\textbf{86.27\%}}  & \textbf{4.07}          \\ \midrule
\multicolumn{7}{c}{\textbf{Chitchat Topics}}                                                                                                                                                                                                                                                                 \\ \midrule
\textbf{PLATO}      & \multicolumn{1}{c|}{\textbf{no knowledge}}                                                & 81.25\%             & 89.29\%                & \textbf{5.36\%}                                                                  & \multicolumn{1}{c|}{66.07\%}           & 3.00                   \\
\textbf{PLATO-SINC} & \multicolumn{1}{c|}{\textbf{web top 1}}                                                   & 81.48\%             & 97.22\%                & 8.17\%                                                                  & \multicolumn{1}{c|}{83.80\%}           & 3.92                   \\
\textbf{PLATO-SINC} & \multicolumn{1}{c|}{\textbf{service}}                                                     & \textbf{83.64\%}    & \textbf{98.60\%}       & 7.94\%                                                        & \multicolumn{1}{c|}{\textbf{84.58\%}}  & \textbf{3.93}          \\ \midrule
\multicolumn{7}{c}{\textbf{In-depth Topics}}                                                                                                                                                                                                                                                                   \\ \midrule
\textbf{PLATO}      & \multicolumn{1}{c|}{\textbf{no knowledge}}                                                & 75.64\%             & 90.60\%                & 16.24\%                                                                 & \multicolumn{1}{c|}{61.54\%}           & 2.58                   \\
\textbf{PLATO-SINC} & \multicolumn{1}{c|}{\textbf{web top 1}}                                                   & 86.32\%             & 97.58\%                & 12.74\%                                                                 & \multicolumn{1}{c|}{81.13\%}           & 3.65                   \\
\textbf{PLATO-SINC} & \multicolumn{1}{c|}{\textbf{service}}                                                     & \textbf{87.38\%}    & \textbf{98.13\%}       & \textbf{7.01\%}                                                         & \multicolumn{1}{c|}{\textbf{86.45\%}}  & \textbf{4.13}          \\ \midrule
\multicolumn{7}{c}{\textbf{Spatiotemporal \& Skill Topics}}                                                                                                                                                                                                                                               \\ \midrule
\textbf{PLATO}      & \multicolumn{1}{c|}{\textbf{no knowledge}}                                                & 69.34\%             & 78.77\%                & 24.06\%                                                                 & \multicolumn{1}{c|}{49.06\%}           & 2.00                   \\
\textbf{PLATO-SINC} & \multicolumn{1}{c|}{\textbf{web top 1}}                                                   & 82.76\%             & 97.57\%                & 22.41\%                                                                 & \multicolumn{1}{c|}{70.69\%}           & 2.98                   \\
\textbf{PLATO-SINC} & \multicolumn{1}{c|}{\textbf{service}}                                                     & \textbf{91.26\%}    & \textbf{98.54\%}       & \textbf{7.77\%}                                                         & \multicolumn{1}{c|}{\textbf{87.86\%}}  & \textbf{4.16}          \\ \bottomrule
\end{tabular}
\caption{Human evaluation results for different types of topics.}
\label{tab:Humanmetric}
\end{table*}

\subsection{Human Evaluation Results}
As demonstrated in Table \ref{tab:Humanmetric}, the model PLATO-SINC, which uses service information, is significantly improved compared to the methods that do not link the world through web information. Specifically, compared to PLATO, the dialogue consistency increases by 11.91\%, the knowledgeable increases by 11.83\% to 98.30\%, and the ratio of factually incorrect decreases by 7.43\%. It is worth mentioning that the engaging significantly improved by 27.42\%, the session-level overall score from 2.53 to 4.07, a relative improvement of 60.87\%. In our opinion, it makes responses more interesting by adding dynamic spatiotemporal information. Compared with the method of using web knowledge, our method is also more advantageous in various dimensions, which benefits from the service information providing broader and more accurate knowledge. 

We further analyze the performance of each method under different types of dialogue topics. On the topic of chitchat, PLATO-SINC showed a slight improvement in consistency and factual accuracy. We consider that the completion of such topics require little external information. The increase in knowledge leads to a decrease in factual accuracy, but still a significant increase in engaging and overall score. A certain amount of external information is required on in-depth topics to complete the conversation. The impact of the PLATO-SINC approach employing service information has significantly improved, mainly based on 98.13\% knowledgeable, where the factual error is only 7.01\%. On the topic of spatiotemporal sensitivity and skills, service information shows its powerful capabilities. The factually incorrect of our method is only 7.77\%, when the other methods are at least 22.41\%. Sensing time and location and possessing comprehensive abilities effectively is beneficial for enhancing conversational engaging. The overall score improved to 4.16, which is unprecedented. A dialog case on information seeking and response generation is in Appendix \ref{sec:case}, and we will add more cases later.

\section{Related Work}
State-of-the-art open-domain dialogue systems are usually implemented by training end-to-end generative models on large amounts of dialog corpus. They generate responses using knowledge learned from static training data, frozen in model weights (\citealp{adiwardana2020towards}; \citealp{roller2020recipes}; \citealp{bao2021plato}; \citealp{gu2022eva2}\label{1}). Many studies in recent years have found that external information can improve the knowledgeability of dialogue responses and reduce the rate of hallucinations (\citealp{zhou2021earl}; \citealp{meng2020dukenet}; \citealp{kim2020sequential}; \citealp{zhou2018dataset}). TopicalChat \cite{gopalakrishnan2019topical}, DuConv \cite{wu2019proactive}, KdConv \cite{zhou2020kdconv}, PersonaChat, etc., provide conversation-related information such as textual knowledge or user portraits, and analyze whether chatbots can use them in conversations. However, these static sources of information are difficult to cover the world knowledge in the open field. And they do not focus on how to retrieve this information, which is important for open-domain dialogue.

Some work attempts to get chatbots to retrieve external information in conversations. A representative example is Wikipedia of Wizard, which uses conversational context to match relevant knowledge documents from Wikipedia \cite{dinan2018wizard}. LaMDA can generate queries to access a question-and-answer database as well as calculator and translation skills to generate responses \cite{thoppilan2022lamda}. The work that is closer to us is Wikipedia of Internet, which obtains relevant web page texts by accessing search engines, expands the scope of information sources, improves the timeliness and coverage of information, and verifies that this method can improve dialog informativeness and factual accuracy \cite{komeili2021internet}. On this basis, \citet{shuster2022language} further proposes to condense the retrieved web pages into concise text knowledge to reduce the interference of the noise in the web page text knowledge to the system. Different from these studies, our goal is to build a chatbot that can link to the real world, acquires world knowledge and state in service information, and has the ability to perceive the spatiotemporal scene like a human.

As shown in Table \ref{tab:Comparison}, human-labeled open-domain dialogue datasets based on external information are scarce, especially in Chinese. DuConv is a human-annotated dataset of knowledge dialogues in the film domain, using triple text as an external knowledge source \cite{wu2019proactive}. KdConv expands the domain of knowledge dialogue into three domains: film, music, and travel \cite{zhou2020kdconv}. In contrast to these, DuSinc is the first open-domain Chinese service information-enhanced dialogue dataset, in which the information source is not a static knowledge base, but the world information linked through service system.

\section{Conclusion}
% We propose a method for dialogue systems to access dynamic spatiotemporal awareness knowledge by linking the real world. We use service information as a representation of the world. And a high-quality Chinese dialogue dataset DuSinc is collected to train the model's ability to actively get service information and generate responses. We build a benchmark model that includes two tasks, query generation and response generation, and demonstrate through extensive experiments that our method can significantly improve the performance of open-domain dialogue on consistency, knowledge richness, factual accuracy and Engaging.

We propose a method to improve dialogue systems with dynamic spatiotemporally aware service information. We collect a high-quality Chinese dialogue dataset DuSinc to train the ability of model to actively get service information and generate responses. We build a benchmark model covering two tasks query generation and response generation. We demonstrate through extensive experiments that our method can significantly improve the performance of open-domain dialogue in terms of consistency, informativeness, factuality, and engagingness. Dialogue systems with real-time access to dynamic knowledge and perception of the spatiotemporal state behave more like humans.

% In future work, we consider continuously expanding the number of conversations in DuSinc and the types of service information. At the same time, we hope to explore the introduction of Task-oriented dialog capabilities and multi-modal resources into open-domain dialogue generation by linking the world.

In future work, we consider expanding the kinds of service information in DuSinc. Exploring the introduction of cross-genre, cross-modal information into open-domain conversations through linking the world.

% \section{Limitations}
% The PLATO-SINC model also faces the same problem as the current language model. The dialogue responses generated by the model can follow the knowledge provided by the service to a certain extent, but the hallucination problem cannot be avoided. At the same time, it is also challenging for the model to find whether the queried knowledge is irrelevant to the dialogue. Using irrelevant knowledge may lead to the loss of dialogue consistency. To ensure the generality of the system, all the services we aggregated receive natural language queries, which may make it difficult for some specific forms of query services to access the system directly.

\section{Ethical Considerations}
This section addresses strategies and techniques for schools, laboratories, corporations, etc., to utilize our work ethically.
First and foremost, we will only collect the spatiotemporal data of user with their permission.
We would release the DuSinc dataset and PLATO-SINC model under CC-BY-SA to prevent illegal usage. Second, DuSinc and PLATO-SINC should identify incorrect wording.
Professional collectors will delete abusive remarks from DuSinc.
We will use profanity dictionaries and algorithmic approaches to detect problematic utterances. Children and teenagers will benefit from more appropriate language.
We will design safeguards and tactics to make PLATO-SINC safe and ethical. Automated systems will groom PLATO-SINC and its users. Using and maintaining session data is regulated by law.

\bibliography{tacl2021}
\bibliographystyle{acl_natbib}

% \iftaclpubformat

\onecolumn
% \clearpage
\appendix
\section{Data Collection Details}
\label{sec:data}
\begin{figure}[!h]
    \centering
    \subfigure[USER side]{
    \includegraphics[width=7.5cm]{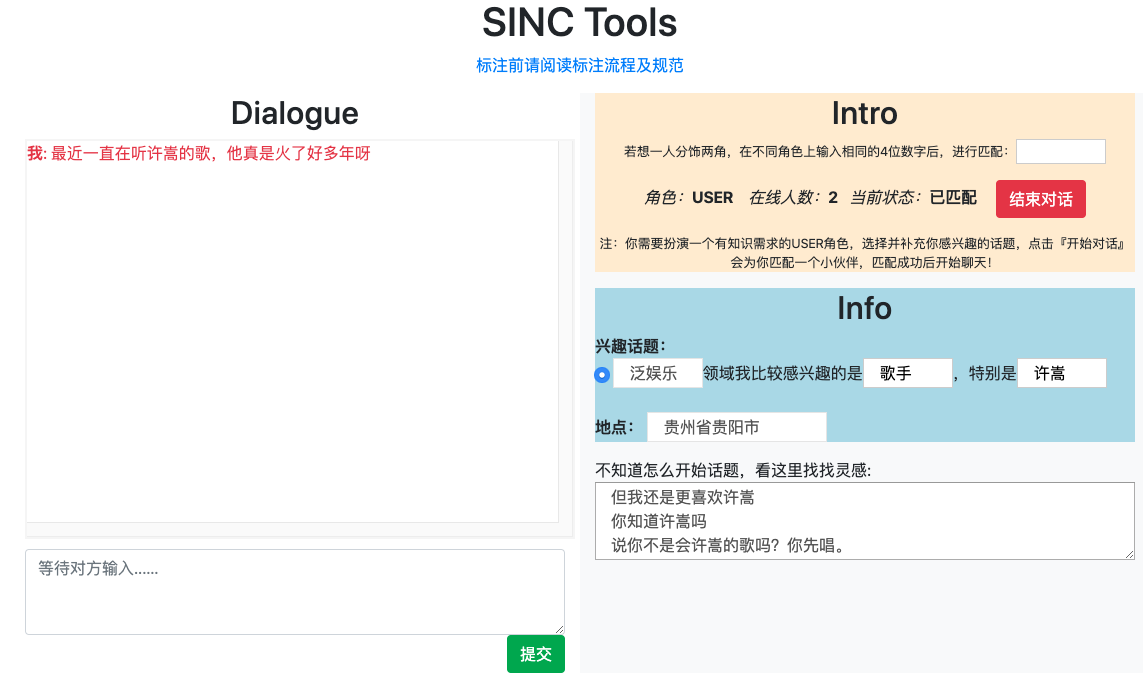}
    }\subfigure[BOT side]{
    \includegraphics[width=7.5cm]{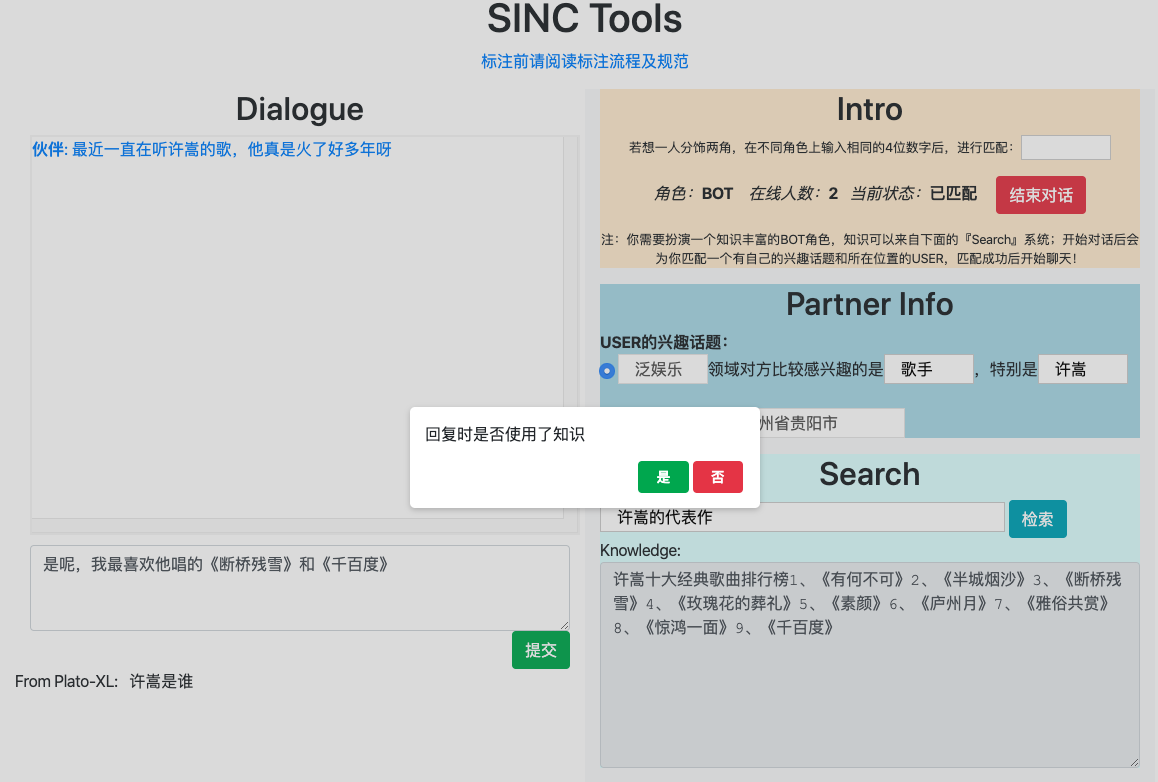}
    }
    \caption{Screenshots of DuSinc data collection tool.}
    \label{fig:screen}
\end{figure}

We developed a data collection tool, which is different on the USER side and the BOT side, Figures \ref{fig:screen} is the corresponding screenshot. The USER will enter the matching state after completing the topic selection, and the BOT can directly enter the matching state. Two different roles that are in the matching state will be matched together. At this time, the topic and location information of the USER will be synchronized to the BOT. The BOT side requires the annotator to choose whether to use the knowledge when submitting the reply, and the system will judge whether the conversation replies highly duplicates the knowledge.

\section{Case Study}
\label{sec:case}

\begin{table*}[!h]\footnotesize
\centering
\begin{tabular}{lll}
\hline
\multicolumn{3}{c}{Spatiotemporal state(\textbf{Time:} 2022.08.12 15:00; \textbf{Location:} Haidian district, Beijing, China )}\\
\hline
\multicolumn{3}{c}{\textbf{Context:} }\\
\multicolumn{3}{c}{\makecell[c]{ user: It's the weekend soon, I'm going to go on an outing with my friends. \begin{CJK*}{UTF8}{gkai} 马上到周末了，我打算和朋友去郊游\end{CJK*} }}\\
\multicolumn{3}{c}{\makecell[c]{bot: That's great, enjoy your vacation. 
\begin{CJK*}{UTF8}{gkai} 那很棒呀，好好享受假期时光。\end{CJK*}}}\\
\multicolumn{3}{c}{\makecell[c]{user: Hope for good weather.
\begin{CJK*}{UTF8}{gkai} 希望能有个好天气。\end{CJK*}}}\\
% \multicolumn{3}{c}{A: Truthfully, anything with cheese is the best}\\
\hline
\multicolumn{3}{c}{\textbf{Request:} query generation: weekend weather; Latitude: ***; Longitude: ***}\\
\hline
& \textbf{Knowledge Seeking} & \textbf{Response Generation}\\
\hline
service knowledge&\makecell[l]{ In the next 2 days in Beijing, the weather will be \\mainly cloudy, tomorrow will be 18 degrees to \\26 degrees, cloudy; the day after tomorrow, 16 \\degrees to 21 degrees, with light rain. \\
\begin{CJK*}{UTF8}{gkai} \makecell[l]{ 北京未来2天内，天气以多云为主，明天18度\\～26度，多云;后天16度～21度，小雨。} \end{CJK*}}& \makecell[l]{The weather in Beijing will be cloudy tomorrow, \\suitable for going out to play, there will be light\\ rain the day after tomorrow.\\
\begin{CJK*}{UTF8}{gkai} 北京明天天气多云，适合出去玩，后天有小雨。\end{CJK*}}\\
% & of a yeasted flatbread typically topped with & consisting of yeasted flatbread typically topped\\
% & tomato sauce and cheese and baked in an & with tomato sauce and cheese\\
% & oven. & \\
% & (from “Pizza” Wikipedia page)& \\
\hline 
no knowledge & no\_passages\_used &Good weather, good mood.\\
& & \begin{CJK*}{UTF8}{gkai}天气好，心情好。\end{CJK*}\\
\hline
web knowledge & \makecell[l]{
% It is expected that there will be showers or\\
% thunderstorms between cloudy and cloudy\\ days from Friday night to Saturday daytime,\\ with 8 to 15 degrees Celsius.  \\
...... Cloudy and cloudy Saturday night to Sunday \\day. From Sunday evening to night, it will turn to \\showers or thunderstorms, accompanied by local hail, \\strong winds, short-term heavy precipitation and other\\ strong convective weather.  In rainy and foggy weather, \\ please pay attention to traffic safety. \\
% On the weekends, there may be strong convective weather such\\
% as strong winds, hail, and short-term heavy\\ precipitation. Please take precautions.  \\
\textcolor{red}{(No spatiotemporal awareness)} \\
\begin{CJK*}{UTF8}{gkai} \makecell[l]{ 
% 预计周五夜间至周六白天多云间阴天有阵雨或\\
% 雷雨，8～15℃。 \\
...... 周六夜间到周日白天多云间阴天。周日傍晚\\至夜间转阵雨或雷雨，局地伴
有冰雹、大风\\、短时强降水等强对流天气。雨
雾天气\\，出行请注意交通安全。\\
% 周末局地可能伴有大风、冰雹、短时强\\
% 降水等强对流天气，请注意防范。\\
\textcolor{red}{（无时空感知）}}\end{CJK*}
}& \makecell[l]{There will be showers during the day on \\Saturday, please pay attention to traffic\\ safety.\\
\textcolor{red}{(Facts are not accurate)}\\
\begin{CJK*}{UTF8}{gkai} \makecell[l]{周六白天有阵雨，出行请注意交通安全。\\
\textcolor{red}{（事实不准确）}}\end{CJK*}}\\
% \begin{CJK*}{UTF8}{gkai} 周六白天有阵雨，出行请注意交通安全（事实不准确）\end{CJK*}
% }
% & be made without cheese. & \\
% & (from “Pizza” Wikipedia page) &\\
\hline
% Human & It is the second most popular cheese in the & especially cheddar cheese! it's the second most\\
% & US (behind mozzarella), with an average & popular cheese in the use !\\
% & annual consumption of 10 lb (4.5 kg) per & \\
% & capita.&\\
% & (from “Cheddar cheese” Wikipedia page)&\\
% \hline
\end{tabular}
\caption{Cases of seek knowledge and generate responses by our model and baselines.}
\label{tab:case}
\end{table*}

% Figures \ref{fig:case1} and \ref{fig:case2} are dialog examples for service information enhancement.

\end{document}